\documentstyle[colacl]{article}

\title{An Estimate of Referent of Noun Phrases in Japanese Sentences}
\author{\hspace*{-.5cm}Masaki Murata \\
  \hspace*{-.5cm} Communications Research Laboratory\\
  \hspace*{-.5cm}\normalsize 588-2, Iwaoka, Nishi-ku, Kobe, 651-2401, Japan
  \And
\hspace*{.5cm}  Makoto Nagao \\
\hspace*{.5cm}    Kyoto University \\
\hspace*{.5cm}  \normalsize   Yoshida-Honmachi, Sakyo, Kyoto 606-01, Japan}

\begin{document}
\bibliographystyle{fullname}

\baselineskip=1\baselineskip

\maketitle
\begin{abstract}

In machine translation and man-machine dialogue,  
it is important to clarify referents of noun phrases. 
We present a method for determining the referents of noun phrases 
in Japanese sentences 
by using the referential properties, modifiers, 
and possessors\footnote{The possessor of a noun phrase 
is defined as the entity 
which is the owner of the entity denoted by the noun phrase.}
 of noun phrases. 
Since the Japanese language has no articles,  
it is difficult 
to decide whether a noun phrase has an antecedent or not. 
We had previously estimated 
the referential properties of noun phrases 
that correspond to articles 
by using clue words in the sentences \cite{match}. 
By using these referential properties, 
our system determined the referents of noun phrases in Japanese sentences. 
Furthermore we used the modifiers and possessors of noun phrases 
in determining the referents of noun phrases. 
As a result, on training sentences 
we obtained a precision rate of 82\% 
and a recall rate of 85\% in the determination 
of the referents of noun phrases that have antecedents.  
On test sentences, 
we obtained a precision rate of 79\% 
and a recall rate of 77\%. 
\end{abstract}
\section{Introduction}
\label{sec:intro}

This paper describes the determination of 
the referent of a noun phrase 
in Japanese sentences. 
In machine translation, 
it is important to clarify 
the referents of noun phrases. 
For example, 
since the two ``OJIISAN (old man)'' in the following sentences 
have the same referent,
the second ``OJIISAN (old man)'' should be pronominalized 
in the translation into English. 
\begin{equation}
  \begin{minipage}[h]{10.5cm}
\small
    \begin{tabular}[t]{l@{ }l@{ }l}
\underline{OJIISAN}-WA & JIMEN-NI & KOSHI-WO-OROSHITA.\\
(old man) & (ground) & (sit down)\\
\multicolumn{3}{l}{
(\underline{The old man} sat down on the ground.)}\\[0.3cm]
\end{tabular}
\begin{tabular}[t]{l@{ }l@{ }l}
YAGATE & \underline{OJIISAN}-WA & NEMUTTE-SHIMATTA.\\
(soon) & (old man) & (fall asleep)\\
\multicolumn{3}{l}{
(\underline{He (= the old man)} soon fell asleep.)}\\
\end{tabular}\\
  \end{minipage}
\label{eqn:ojiisan_jimen_meishi}
\end{equation}
When dealing with a situation like this, 
it is necessary for a machine translation system to recognize that 
the two ``OJIISAN (old man)'' have the same referent. 
In this paper, 
we propose a method that determines the referents of noun phrases 
by using (1) the referential properties of noun phrases, 
(2) the modifiers in noun phrases,
 and (3) the possessors
of entities denoted by the noun phrases. 

For languages that have articles, like English, 
we can use articles (``the'', ``a'', and so on) 
to decide whether a noun phrase has an antecedent or not. 
In contrast, 
for languages that have no articles, like Japanese, 
it is difficult to decide 
whether a noun phrase has an antecedent. 
We previously estimated 
the referential properties of noun phrases 
that correspond to articles 
for the translation of Japanese noun phrases into English \cite{match}.
By using these referential properties, 
our system determines the referents of noun phrases 
in Japanese sentences. 
Noun phrases are classified by referential property into 
generic noun phrases, 
definite noun phrases, 
and indefinite noun phrases. 
When the referential property of a noun phrase is a definite noun phrase, 
the noun phrase can refer to the entity 
denoted by a noun phrase that has already appeared. 
When the referential property of a noun phrase 
is an indefinite noun phrase 
or a generic noun phrase, 
the noun phrase cannot refer to the entity 
denoted by a noun phrase that has already appeared. 

It is insufficient to determine referents of noun phrases 
using only the referential property. 
This is because 
even if the referential property of a noun phrase is a definite noun phrase, 
the noun phrase does not refer to the entity denoted by a noun phrase 
which has a different modifier or possessor. 
Therefore, we also use the modifiers and possessors of noun phrases 
in determining referents of noun phrases. 

In connection with our approach, 
we would like to emphasize the following points: 
\begin{itemize}
\item 
So far little work has been done 
on determining the referents of noun phrases in Japanese. 

\item 
Since the Japanese language has no articles, it is difficult 
to decide whether a noun phrase has an antecedent or not. 
We use referential properties to solve this problem. 

\item 
We determine the possessors of entities denoted by noun phrases and 
use them like modifiers in estimating the referents of noun phrases. 
Since the method uses the sematic relation between 
an entity and the possessor, which is a language-independent knowledge, 
it can be used in any other language. 
\end{itemize}

\section{Referential Property of a Noun Phrase}

The following is an example of noun phrase anaphora. 
``OJIISAN (old man)'' in the first sentence and 
``OJIISAN (old man)'' in the second sentence 
refer to the same old man, 
and they are in anaphoric relation. 
\begin{equation}
  \begin{minipage}[h]{11.5cm}
    \small
    \begin{tabular}[t]{l@{ }l@{ }l@{ }l@{ }l}
    \underline{OJIISAN} & TO & OBAASAN-GA & SUNDEITA.\\
    (an old man)&(and)&(an old woman)&(lived)\\
\multicolumn{5}{l}{
  (There lived \underline{an old man} and an old woman.)}
    \end{tabular}
\vspace{0.3cm}

    \begin{tabular}[t]{l@{ }l@{ }l@{ }l}
      \underline{OJIISAN}-WA & YAMA-HE & SHIBAKARI-NI & ITTA.\\
      (old man)  & (mountain) & (to gather firewood) & (go)\\
\multicolumn{4}{l}{
  (\underline{The old man} went to the mountains to gather firewood.)}
    \end{tabular}

  \end{minipage}
\label{eqn:ojiisan_obaasan_meishi}
\end{equation}
%


When the system analyzes the anaphoric relation of noun phrases like these, 
the referential properties of noun phrases are important. 
The referential property of a noun phrase here means 
how the noun phrase denotes the referent. 
If the system can recognize that 
the second ``OJIISAN (old man)'' 
has the referential property of the definite noun phrase,  
indicating that the noun phrase 
refers to the contextually non-ambiguous entity, 
it will be able to judge that 
the second ``OJIISAN (old man)'' refers to 
the entity denoted by the first ``OJIISAN (old man). 
The referential property plays an important role 
in clarifying the anaphoric relation. 

We previously classified noun phrases 
by referential property 
into the following three types \cite{match}. 

{\scriptsize
\[\rm \mbox{\normalsize NP}
 \left\{ \begin{array}{l}
     \rm \mbox{\normalsize {\bf generic} NP}\\
         \mbox{\normalsize {\bf non generic} NP}
            \left\{ \begin{array}{l}
              \rm \mbox{\normalsize {\bf definite} NP}\\
                  \mbox{\normalsize {\bf indefinite} NP}
            \end{array}
            \right.
\end{array}
\right.
\] 
}

\paragraph{Generic noun phrase}
A noun phrase is classified as generic 
when it denotes all members of the class described by the noun phrase 
or the class itself of the noun phrase. 
For example, ``INU(dog)'' in the following sentence is a generic noun phrase.
\begin{equation}
  \begin{minipage}[h]{11.5cm}
    \small
    \begin{tabular}[t]{ll}
  \underline{INU-WA} & YAKUNI-TATSU.\\
  (dog) & (useful)\\
\multicolumn{2}{l}{
  (\underline{Dogs} are useful.)}
    \end{tabular}
  \end{minipage}
  \label{eqn:3c_doguse}
\end{equation}
A generic noun phrase cannot refer to the entity 
denoted by an indefinite or definite noun phrase. 
Two generic noun phrases can have the same referent. 
%
\paragraph{Definite noun phrase}
A noun phrase is classified as definite 
when it denotes a contextually non-ambiguous member 
of the class of the noun phrase. 
For example, ``INU(dog)'' 
in the following sentence is a definite noun phrase.
\begin{equation}
  \begin{minipage}[h]{11.5cm}
    \small
    \begin{tabular}[t]{lll}
  \underline{INU-WA} & MUKOUHE & ITTA.\\
  (dog) & (away) & (go)\\
\multicolumn{3}{l}{
  (\underline{The dog} went away.)}
\end{tabular}
\end{minipage}
\label{eqn:3c_thedoguse}
\end{equation}
A definite noun phrase can refer to the entity denoted by a noun phrase 
that has already appeared. 
\paragraph{Indefinite noun phrase}
An indefinite noun phrase denotes 
an arbitrary member of the class of the noun phrase.
For example, ``INU(dog)'' in the following sentence 
is an indefinite noun phrase.
\begin{equation}
  \begin{minipage}[h]{11.5cm}
    \small
    \begin{tabular}[t]{lll}
    \underline{INU-GA} & SANBIKI & IRU.\\    
    (dog) & (three) & (there is)\\
\multicolumn{3}{l}{
  (There are three \underline{dogs}.)}
\end{tabular}
\end{minipage}
  \label{eqn:3_threedogs}
\end{equation}
An indefinite noun phrase cannot refer to 
the entity denoted by  a noun phrase 
that has already appeared. 

\section{How to Determine the Referent of a Noun Phrase}
\label{sec:how_to}

To determine referents of noun phrases, 
we made the following three constraints. 
\begin{enumerate}
\item 
Referential property constraint 
\item 
Modifier constraint 
\item 
Possessor constraint 
\end{enumerate}
When two noun phrases 
which have the same head noun 
satisfy these three constraints, 
the system judges that the two noun phrases have the same referent. 

\subsection{Referential Property Constraint}
\label{sec:ref_pro}

First, our system estimates the referential property of a noun phrase
by using the method described in one of our previous papers \cite{match}. 
The method estimates a referential property 
using surface expressions in the sentences. 
For example, 
since the second ``OJIISAN (old man)'' 
in the following sentences 
is accompanied by a particle ``WA (topic)''  
and the predicate is in the past tense, 
it is estimated to be a definite noun phrase. 
\begin{equation}
  \begin{minipage}[h]{11.5cm}
    \small
\vspace{0.3cm}
    \begin{tabular}[t]{l@{ }l@{ }l}
\underline{OJIISAN}-WA & JIMEN-NI &
KOSHI-WO-OROSHITA.\\
(old man) & (ground) & (sit down)\\ 
\multicolumn{3}{l}{
(\underline{The old man} sat down on the ground.)}\\[0.3cm]
\end{tabular}
\begin{tabular}[t]{l@{ }l@{ }l}
YAGATE & \underline{OJIISAN}-WA & 
NEMUTTE-SHIMAIMATTA.\\
(soon) & (old man) & (fall asleep)\\
\multicolumn{3}{l}{
(\underline{He} soon fell asleep.)}
\end{tabular}
  \end{minipage}
\label{eqn:ojiisan_jimen_meishi_2}
\end{equation}

Next, our system determines the referent of a noun phrase 
by using its estimated referential property. 
When a noun phrase is estimated to be a definite noun phrase, 
our system judges that the noun phrase refers to 
the entity denoted by a previous noun phrase 
which has the same head noun. 
For example, 
the second ``OJIISAN'' in the above sentences 
is estimated to be a definite noun phrase, 
and our system judges that it refers to 
the entity denoted by the first ``OJIISAN''. 

When a noun phrase is not estimated to be a definite noun phrase, 
it usually does not refer to the entity denoted by 
a noun phrase that has already been mentioned. 
Our method, however, might fail to estimate the referential property,  
so the noun phrase might refer to the entity denoted by 
a noun phrase that has already been mentioned.  
Therefore, 
when a noun phrase is not estimated to be a definite noun phrase, 
our system gets a possible referent of the noun phrase 
and determines whether or not the noun phrase refers to it 
by using the following three kinds of information. 
\begin{itemize}
\item \underline{the plausibility($P$) 
of the estimated referential} \underline{property
that is a definite noun phrase}\\[0.1cm]
When our system estimates a referential property, 
it outputs the score of each category \cite{match}. 
The value of the plausibility ($P$) is 
given by the score.

\item
\underline{the weight ($W$) of the salience of a possible}
\underline{referent}\\[0.1cm]
The weight ($W$) of the salience is given 
by the particles such as "WA ({\sf topic})" and "GA ({\sf subject})". 
The entity denoted by a noun phrase 
which has a high salience, 
is easy to be referred by a noun phrase. 

\item
\underline{the distance ($D$)
between the estimated noun} 
\underline{phrase and 
a possible referent}\\[0.1cm]
The distance ($D$) is the number of noun phrases 
between the estimated noun phrase and 
a possible referent. 
\end{itemize}
When the value given by these three kinds of information 
is higher than a given threshold, our system 
judges that the noun phrase refers to the possible referent. 
Otherwise, it judges that the noun phrase does not refer to the possible
referent and is an indefinite noun phrase or a generic noun phrase. 

\subsection{Modifier Constraint}

It is insufficient to determine referents of noun phrases 
by using only the referential property. 
When two noun phrases have different modifiers, 
they usually do not have the same referent. 
For example, 
``MIGI(right)-NO HOO(cheek)'' and 
``HIDARI(left)-NO HOO(cheek)'' in the following sentences 
do not have the same referent.  
\begin{equation}
  \begin{minipage}[h]{11.5cm}
\small
\vspace{0.3cm}
\hspace*{-0.6cm}
    \begin{tabular}[t]{l@{ }l@{ }l@{ }l@{ }l@{ }l}
{\footnotesize KONO} & {\footnotesize OJIISAN-NO} & {\footnotesize KOBU-WA} & 
{\footnotesize \underline{MIGI-NO}} & {\footnotesize \underline{HOO}-NI} & {\footnotesize ATTA.}\\
(this) & (old man) & (lump) & (right)  & (cheek) & (be on)\\
\multicolumn{6}{l}{
(This old man's lump was on \underline{his right cheek}.)}\\[0.3cm]
\end{tabular}

\hspace*{-0.6cm}
\begin{tabular}[t]{l@{ }l@{ }l@{ }l@{ }l}
{\footnotesize TENGU-WA,} & {\footnotesize KOBU-WO} &
{\footnotesize \underline{HIDARI-NO}} & {\footnotesize \underline{HOO}-NI} & 
{\footnotesize TSUKETA.}\\
(tengu)\footnotemark & (lump) & (left) & (cheek) & (put on)\\
\multicolumn{5}{l}{
(The "tengu" put a lump on \underline{his left cheek})}
\end{tabular}
  \end{minipage}
\label{eqn:mouhitori_ojiisan_hoho_kobu}
\end{equation}
\footnotetext{
A tengu is a kind of monster.} 
Therefore, we made the following constraint: 
A noun phrase that has a modifier cannot refer to the entity 
denoted by a noun phrase that does not have the same modifier. 
A noun phrase that does not have a modifier 
can refer to the entity denoted by a noun phrase that has any modifier. 

The constraint is incomplete, 
and is not truly applicable to all cases. 
There are some exceptions where 
a noun can refer to the entity of a noun that has a different modifier. 
But we use the constraint because 
we can get a higher precision than if we did not use it. 

\subsection{Possessor Constraint}

When a noun phrase has a semantic marker 
{\sf PAR} (a part of a body),\footnote{
In this paper, we use the Noun Semantic Marker Dictionary
\cite{imiso-in-BGH}. 
}  
our system tries to estimate 
the possessor of the entity denoted by the noun phrase. 
We suppose that 
the possessor of a noun phrase is the subject or 
the noun phrase's nearest topic 
that has a semantic marker {\sf HUM} (human) or 
a semantic marker {\sf ANI} (animal). 
For example, 
we examine two instances of ``HOO (cheek)'' in the following sentences, 
which have a semantic marker {\sf PAR}. 

\vspace{0.3cm}

{\small
    \begin{tabular}[t]{l@{ }l@{ }l}
      OJIISAN-NIWA & \underline{[OJIISAN-NO]\footnotemark
} &
    \underline{HIDARI-NO} \\
    (old man) & (old man's) & (left) \\
\end{tabular}

    \begin{tabular}[t]{l@{ }l@{ }l}
      \underline{HOO-NI} & KOBU-GA & ATTA.\\
      (cheek) & (lump) & (be on)\\
\multicolumn{3}{l}{
  (This old man had a lump on \underline{his left cheek}.)} \\[0.3cm]
\end{tabular}
\footnotetext{
\label{foot:bracket}
The words in brackets [ ] are omitted in the sentences.} 

\begin{tabular}[t]{l@{ }l@{ }l@{ }l}
    SORE-WA & KOBUSHI-HODO-NO & KOBU-DATTA.\\
    (it) &  (person's fist) & (lump)\\
\multicolumn{4}{l}{
  (It is about the size of a person's fist.)}\\[0.3cm]
\end{tabular}

\begin{tabular}[t]{l@{ }ll}
  OJIISAN-GA & 
  \underline{[OJIISAN-NO]}
 & \underline{HOO-WO} \\
    (old man ({\sf subject})) & (old man's) & (cheek) \\
  \end{tabular}\\

\begin{tabular}[t]{l@{ }l}
 HUKURAMASETE &
 IRUYOUNI-MIETA.\\
  (puff) & (look as if)\\
    \multicolumn{2}{l}{
      (He looked as if he had puffed out \underline{his cheek}.)}
  \end{tabular}\\
}

\vspace{0.3cm}

\noindent
The possessor of the first ``HOO (cheek)'' 
is determined to be ``OJIISAN (old man)'' 
because ``OJIISAN (old man)'', 
which has a semantic marker {\sf HUM} (human),  
is followed by a particle ``NIWA ({\sf topic})'' 
and is the topic of the sentence.  
The possessor of the second ``HOO (cheek)'' 
is also determined to be ``OJIISAN (old man)'' 
because 
``OJIISAN (old man)'' is the subject of the sentence. 

We made the following constraint, 
which is similar to the modifier constraint, 
by using possessors.  
When the possessor of a noun phrase is estimated, 
the noun phrase cannot refer to the entity denoted by 
a noun phrase that does not have the same possessor. 
When the possessor of a noun phrase is not estimated, 
the noun phrase can refer to the entity denoted by 
a noun phrase that has any possessor. 

For example, 
since the two instances of ``HOO (cheek)'' in the above sentences 
have the same possessor ``OJIISAN (old man)'', 
our system correctly judges that they 
have the same referent. 

\section{Anaphora Resolution System}

\subsection{Procedure}
\label{wakugumi}

Before referents are determined, 
sentences are transformed into a case structure 
by the case structure analyzer \cite{kuro}. 

\begin{table*}[t]

\small

  \fbox{
  \begin{minipage}[h]{15cm}
  \begin{center}

\caption{Results}
  \label{tab:noun_result}

\vspace{0.3cm}

    \leavevmode
    \newcommand{\mn}[1]{\begin{minipage}[t]{3cm}%
      \baselineskip0.8cm%
      #1 \smallskip \end{minipage}}
\begin{tabular}[c]{|l|r@{ }c|r@{ }c|}\hline
 &  \multicolumn{2}{|c|}{{Precision}}& \multicolumn{2}{|c|}{{Recall}}\\\hline
Training sentences      &  82\% & (130/159) & 85\% & (130/153) \\
Test sentences &  79\% & ( 89/113) &  77\% & ( 89/115)\\\hline
\end{tabular}
  \end{center}
Training sentences \{example sentences (43 sentences), a folk tale ``KOBUTORI
JIISAN'' \cite{kobu} (93 sentences), an essay in ``TENSEIJINGO'' (26 sentences), an editorial (26 sentences), an article in ``Scientific American (in Japanese)''(16 sentences)\}\\
Test sentences \{a fork tale ``TSURU NO ONGAESHI'' \cite{kobu} (91 sentences), two essays in ``TENSEIJINGO'' (50 sentences), an editorial (30 sentences), ``Scientific American(in Japanese)'' (13 sentences)\}
  \end{minipage}
  }
\end{table*}

\begin{table*}[t]

\small

  \fbox{
  \begin{minipage}[h]{15cm}
  \begin{center}
\caption{Comparison}
  \label{tab:sijisei_taishou}

\vspace{0.3cm}

    \leavevmode
    \newcommand{\mn}[1]{\begin{minipage}[t]{3cm}%
      \baselineskip0.8cm%
      #1 \smallskip \end{minipage}}
\begin{tabular}[c]{|ll|r@{ }c|r@{ }c|r@{ }c|r@{ }c|}\hline
& &
 \multicolumn{2}{|c|}{Method 1} & 
 \multicolumn{2}{c|}{Method 2} & 
 \multicolumn{2}{c|}{Method 3} & 
 \multicolumn{2}{c|}{Method 4}\\\hline 
Training sentences & \multicolumn{1}{|l|}{Precision}    &  82\% & (130/159) &  92\% & (117/127) &  72\% & (123/170) &  65\% & (138/213) \\
                   & \multicolumn{1}{|l|}{Recall}       &  85\% & (130/153) &  76\% & (117/153) &  80\% & (123/153) &  90\% & (138/153) \\\hline
Test sentences     & \multicolumn{1}{|l|}{Precision}    &  79\% & ( 89/113) &  92\% & ( 78/ 85) &  69\% & ( 79/114) &  58\% & ( 92/159) \\
                   & \multicolumn{1}{|l|}{Recall}       &  77\% & ( 89/115) &  68\% & ( 78/115) &  69\% & ( 79/115) &  80\% & ( 92/115) \\\hline
\end{tabular}
  \end{center}
{
Method 1 : The method used in this work

Method 2 : Only when it is estimated to be definite can it refer to the entity
denoted by a noun phrase

Method 3 : No use of referential property

Method 4 : No use of modifier constraint and possessor constraint

}
\end{minipage}
}
\end{table*}

Referents of noun phrases are 
determined by using heuristic rules which are 
made from information such as the three constraints 
mentioned in Section~\ref{sec:how_to}. 
Using these rules, our system takes possible referents 
and gives them points. It judges that 
the candidate having the maximum total score is the referent. 
This is because a number of types of information are combined 
in anaphora resolution. 
We can specify 
which rule takes priority 
by using points. 

The heuristic rules 
are given in the following form. 

\vspace{-0.3cm}

\begin{center}
    \begin{minipage}[c]{8cm}
      \hspace*{0.7cm}{\sl Condition} $\Rightarrow$ \{ Proposal Proposal .. \}\\
      \hspace*{0.7cm}Proposal := ( {\sl Possible-Referent} \, {\sl Point} )
    \end{minipage}
\end{center}

\noindent
Here, {\sl Condition} consists of 
surface expressions, semantic constraints and 
referential properties. 
In {\sl Possible-Referent}, 
a possible referent, 
``Indefinite'', 
``Generic'', 
or other things are written. 
``Indefinite'' means that 
the noun phase is an indefinite noun phrase, 
and it does not refer to the entity denoted by a previous noun phrase. 
{\sl Point} means the plausibility value of the possible referent. 

\subsection{Heuristic Rule for Estimating Referents}
\label{sec:3c_rule}

We made 8 heuristic rules for the resolution of noun phrase anaphora.  
Some of them are given below. 

\begin{enumerate}
\item[R1] 
  When a noun phrase is modified by the words ``SOREZORE-NO (each)'' and ``ONOONO-NO (each)'', \\
  \{(Indefinite, \,$25$)\}

\item[R2]  
When a noun phrase is estimated to be a definite noun phrase, 
and satisfies the modifier and possessor constraints, 
and the same noun phrase X has already appeared, 
  \\ \{(The noun phrase X, \,$30$)\}

\item[R3]  
  When a noun phrase is estimated to be a generic noun phrase, \\
  \{(Generic, \,$10$)\} 

\item[R4] 
  When a noun phrase is estimated to be an indefinite noun phrase, \\
  \{(Indefinite, \,$10$)\}

\item[R5]  
  When a noun phrase X is not estimated to be a definite noun phrase, \\
  \{
  (A noun phrase X which satisfies 
  the modifier and possessor constraints, \, $P+W-D+4$)\}\\ 
  The values $P$, $W$, $D$ are as defined in Section \ref{sec:ref_pro}. 

\end{enumerate}

\section{Experiment and Discussion}

\subsection{Experiment}
\label{sec:eval}

Before determining the referents of noun phrases, 
sentences were at first 
transformed into a case structure 
by the case structure analyzer \cite{kuro}. 
The errors made by the case analyzer were corrected by hand. 
Table \ref{tab:noun_result} shows the results of 
determining the referents of noun phrases.  

To confirm that the three constraints
(referential property, modifier, and possessor) are effective, 
we experimented under several different conditions  
and compared them. 
The results are shown in Table~\ref{tab:sijisei_taishou}. 
{\it Precision}\, is 
the fraction  of noun phrases which were judged to have antecedents. 
{\it Recall}\, is the fraction of noun phrases 
 which have antecedents. 

In these experiments we used training sentences and test sentences. 
The training sentences were used to 
make the heuristic rules in Section~\ref{sec:3c_rule} by hand. 
The test sentences 
were used to confirm the effectiveness of these rules. 

In Table \ref{tab:sijisei_taishou}, 
Method~1 is the method mentioned
in Section~\ref{sec:how_to}  
which uses all three constraints.  
Method~2 is the case in which a noun phrase can refer to the entity 
denoted by a noun phrase, 
only when the estimated referential property 
is a definite noun phrase, 
where the modifier and possessor constraints are used. 
Method~3 
does not use a referential property.  
It only uses information such as distance, topic-focus, 
modifier, and possessor. 
Method~4 
does not use the
modifier and possessor constraints. 

The table shows many results. 
In Method~1, 
both the recall and the precision were 
relatively high 
in comparison with the other methods. 
This indicates that 
the referential property was used properly in the method that 
is described in this paper. 
Method~1 was higher than Method~3 
in both recall and precision. 
This indicates that the information of referential property is necessary. 
In Method~2, 
the recall was low 
because 
there were many noun phrases that 
were definite 
but were estimated to be indefinite or generic, and 
the system estimated that 
the noun phrases cannot refer to noun phrases. 
In Method~4, 
the precision was low. 
Since the modifier and possessor constraints were not used, and 
there were many pairs of two noun phrases 
that did not co-refer, such as ``HIDARI(left)-NO HOO(cheek)'' 
and ``MIGI(right)-NO HOO(cheek)'', 
these pairs were incorrectly interpreted to be co-references. 
This indicates that 
it is necessary to use the modifier and possessor constraints. 

\subsection{Examples of Errors}

We found that it was necessary to use modifiers and possessors in 
the experiments. 
But there are some cases 
when the referent was determined incorrectly 
because the possessor of a noun was estimated incorrectly. 

Sometimes a noun can refer to the entity 
denoted by a noun that has a different modifier. 
In such cases, the system made an incorrect judgment. 
\begin{quote}
\small
\hspace*{-0.8cm}
\begin{tabular}[t]{l@{ }l@{ }l@{ }l@{ }l@{ }l@{ }l}
OJIISAN-WA & \underline{CHIKAKU-NO} &
\underline{OOKINA} &  \underline{SUGI-NO}\\
(old man) & (near) &(huge) & (cedar) \\
\end{tabular}

\hspace*{-0.8cm}
\begin{tabular}[t]{l@{ }l@{ }l@{ }l@{ }l@{ }l@{ }l}
\underline{KI-NO} & \underline{NEMOTO-NI}&
\underline{ARU} & \underline{ANA}-DE \\
(tree) & (base) &(be at) & (hole)\\
\end{tabular}

\hspace*{-0.8cm}
\begin{tabular}[t]{l@{ }l}
 AMAYADORI-WO &
SURU-KOTO-NI-SHITA.\\
(take shelter from the rain) & 
(decide to do)\\
\multicolumn{2}{l}{
(So, he decided to take shelter from the rain in \underline{a hole}}\\
\multicolumn{2}{l}{
\underline{which is at the base of a huge cedar tree nearby}.)}\\[0.3cm]
\end{tabular}

(an omission of the middle part)

\vspace{0.3cm}

\hspace*{-0.8cm}
\begin{tabular}[t]{l@{ }l@{ }l@{ }l@{ }l}
{\footnotesize TSUGI-NOHI,} & {\footnotesize KONO} & {\footnotesize OJIISAN-WA} &
{\footnotesize YAMA-HE} & {\footnotesize ITTE,}\\
(next day) & (this) & (old man)& 
(mountain) & (go to)\\
\multicolumn{5}{l}{
(The next day, this man went to the mountain, 
)}\\
\end{tabular}

\vspace{0.3cm}

\hspace*{-0.8cm}
\begin{tabular}[t]{l@{ }l@{ }l@{ }l@{ }l}
{\footnotesize \underline{SUGI-NO}} & {\footnotesize \underline{KI-NO}} & {\footnotesize \underline{NEMOTO-NO}} & {\footnotesize \underline{ANA}-WO} & 
{\footnotesize MITSUKETA.}\\
(cedar) & (tree) & (at base) & (hole) & 
(found)\\
\multicolumn{5}{l}{
(and found \underline{the hole at the base of the cedar tree}.)}\\
\end{tabular}
\end{quote}

\vspace{0.5cm}

The two instances of ``ANA (hole)'' in these sentences refer to the same entity. 
But our system judged that 
they do not refer to it 
because the modifiers of the two instances of ``ANA (hole)'' are different. 
In order to correctly analyze this case, 
it is necessary to decide 
whether the two different expressions are equal in meaning. 

\section{Summary}

This paper describes a method for the determination of referents of 
noun phrases by using their referential properties, 
modifiers, and possessors. 
Using this method on training sentences, 
we obtained a precision rate of 82\% 
and a recall rate of 85\% in the determination 
of referents of noun phrases that have antecedents. 
On test sentences,  
we obtained a precision rate of 79\% 
and a recall rate of 77\%. 
This confirmed 
that 
the use of 
the referential properties, modifiers, and possessors of noun phrases 
is effective.  

\small
\bibliographystyle{plain}

\end{document}